\documentclass{article} 
\usepackage{iclr2021_conference, times}

\usepackage{amsmath,amsfonts,bm}

\def\eqref#1{equation~\ref{#1}}

\def\1{\bm{1}}

\DeclareMathAlphabet{\mathsfit}{\encodingdefault}{\sfdefault}{m}{sl}
\SetMathAlphabet{\mathsfit}{bold}{\encodingdefault}{\sfdefault}{bx}{n}

\def\sN{{\mathbb{N}}}

\def\sR{{\mathbb{R}}}

\usepackage{hyperref}
\usepackage{url}
\usepackage{graphicx}

\title{Investigating Sanity Checks for Saliency Maps with Image and Text Classification}

\author{\hspace{-1cm} Narine Kokhlikyan \hspace{3mm} Vivek Miglani \hspace{3mm} Bilal Alsallakh \hspace{3mm} Miguel Martin \hspace{3mm} Orion Reblitz-Richardson \vspace{2mm}\\
\hspace{5.7cm} Facebook AI \\
\hspace{1.32cm} \texttt{\{narine,vivekm,bilalsal,miguelmartin,orionr\}@fb.com}
}   

\iclrfinalcopy 
\begin{document}

\maketitle

\begin{abstract}
Saliency maps have shown to be both useful and misleading for explaining model predictions especially in the context of images. In this paper, we perform sanity checks for text modality and show that the conclusions made for image do not directly transfer to text. We also analyze the effects of the input multiplier in certain saliency maps using similarity scores, max-sensitivity and infidelity evaluation metrics. Our observations reveal that the input multiplier carries input's structural patterns in explanation maps, thus leading to similar results regardless of the choice of model parameters. We also show that the smoothness of a Neural Network (NN) function can affect the quality of saliency-based explanations. Our investigations reveal that replacing ReLUs with Softplus and MaxPool with smoother variants such as LogSumExp (LSE) can lead to explanations that are more reliable based on the infidelity evaluation metric. 
\end{abstract}

\section{Introduction}
Gradient-based explanation algorithms, also known as saliency maps, help us to explain model predictions~\citep{Ancona2018TowardsBU} w.r.t. input features and internal layers. Despite their popularity, ~\citet{sanity_checks} and ~\citet{sixt2019explanations} show that some gradient-based methods produce explanations that are misleading and invariant to data and parameter randomization. Moreover,~\citet{geometry_is_to_blame} show that the explanations can be easily manipulated by nearly imperceptible perturbations. The latter is tightly connected to the non-smooth nature of explanation functions which suggests that replacing ReLU activations with Softplus increases robustness of explanations.
Although gradient-based methods have been studied thoroughly, their evaluation has been primarily performed for Computer Vision (CV) models, thus their effect on other modalities remains largely unexplored. Accordingly, our contributions are structured as follows:
\begin{itemize}
    \item First, we investigate the applicability of saliency maps in the natural language domain and focus on text classification. We show that saliency maps are more sensitive to parameter randomization when applied to text and less sensitive when applied to image.
    \item Second, we further analyse how the input multiplier affects explanation evaluation metrics and similarity scores using sanity checks both for images and text.
    \item Finally, we show that the smoothness of the NN function is related not only to the quality of model explanations but also to the infidelity~\citep{Yeh2019OnT} evaluation metric. We show that the substitution of ReLU with Softplus and MaxPool with LSE makes explanations not only robust but also more trustworthy based on the infidelity evaluation metric.
\end{itemize}

\section{Related methods and techniques}
~\citet{sanity_checks} propose parameter and data randomization tests, also known as sanity checks, in order to assess the quality of explanation methods.
Parameter randomization implies gradual cascade-like randomization of all model weights starting from the top (logit) moving to the bottom layers of the network. Data randomization implies that explanations are computed for a model that is trained on randomly perturbed ground truth labels.~\citet{sanity_checks} show that most gradient-based explanation methods can be seen as edge detectors since they highlight image contours and are invariant to parameter and data randomization. Based on that~\cite{sanity_checks} claim that gradient-based methods are inadequate tools for model explanation. They also mention that parameter and data invariance is likely to be related to the input multiplier in some gradient-based methods; however, they do not measure input multiplier's quantitative impact.

We show that certain gradient-based methods such as Integrated Gradients (IG), reported to be invariant to parameter and data randomization, do not necessarily act as edge detectors. For the sake of brevity, we report data randomization results in appendices~\ref{sanity_checks_for_image} and~\ref{sanity_checks_for_text}. We perform thorough experiments and show that the observations made for image modality do not necessarily transfer to modalities such as text. We perform our analysis for most explanation methods mentioned in~\citet{sanity_checks}. For brevity, we focus on the IG algorithm because it has both local and global variants~\citep{Ancona2018TowardsBU}. We report experimental results for other methods in Appendix~\ref{sanity_checks_for_image}.
We define an input $\displaystyle x \in \sR^N$, a baseline $\displaystyle x_0 \in \sR^N$, a NN model represented by a continuous function $\displaystyle F : \sR^N \rightarrow \sR^C$, where $C$ is the number of classes and an explanation function $\displaystyle \Phi: \mathcal{F} \times \sR^N \rightarrow \sR^N$. 
Given $\alpha \in [0,1]$ as a scaling factor along the straight line between $x_0$ and $x$, global IG along the $i^{th}$ dimension of the input $x$ is defined as follows:
\begin{equation}
\label{ig_eq}
\Phi({F, x^{i}}) = (x^{i} - x^{i}_0) \cdot \int_0^1\frac{\partial F( x_{0}^{i} + \alpha \cdot (x^{i} - x_{0}^{i}))d\alpha}{dx^{i}}
\end{equation}
The local definition, in contrast, does not include $x^i - x_0^i$ as a multiplier. This multiplier is used to magnify the sensitivities by a constant factor and is independent of model parameters. This explains why global IG fails sanity checks when applied to images~\citep{sanity_checks}. The same observations are valid for InputXGradient~\citep{shrikumar2017just}, DeepLift~\citep{deeplift}, and Gradient SHAP as they incorporate the input multipliers. Certain algorithms such as Guided BackProp and Guided GradCAM, which modify the gradients during back-propagation, do not pass randomization tests for different reasons~\citep{sixt2019explanations}, despite being local.

Since visual evaluation of explanations is subjective and misleading, we compute quantitative evaluation metrics such as infidelity, max-sensitivity and similarity scores between explanation maps.
Infidelity is a generalization of the completeness axiom supported by algorithms such as IG which states that the sum of the explanations is equal to the differences of a NN function at its input and baseline. On the other hand, max-sensitivity measures the sensitivity of model explanations to subtle input perturbations. Lower scores for both metrics indicate more trustworthy explanations.

In terms of similarity metrics, we use the popular Structural Similarity Index (SSIM)~\citep{ssim} for image and Cosine Similarity (CosSim) for text.~\citet{sanity_checks} show that SSIM computed on explanations for original and randomized models retains significant similarity. Therefore we use this metric also for local explanations and compare the results with the global variants. We chose CosSim for text because it is is often used to quantify text similarities. Further metrics could be used instead, such as correlation metrics and Euclidean distance (see Appendix~\ref{sanity_checks_for_text}).

\section{Sanity Checks For Image vs Text }
We compare sanity checks for images and text when model parameters are randomized in cascading manner. We show that local variants of IG for images are sensitive to model parameter randomization while the global variants are not, due to the input multiplier.
In case of text, both global and local IG explanations are sensitive to parameter randomization. Global IG for text is still sensitive to parameter randomization because in contrary to image we sum the explanations along the embedding dimension in order to obtain token-level human interpretable importance scores. The effect of the multipliers becomes insignificant due to small variance and are likely to cancel out in the summation.

\begin{figure}[ht]
\begin{center}
\centerline{\includegraphics[width=1.0\columnwidth]{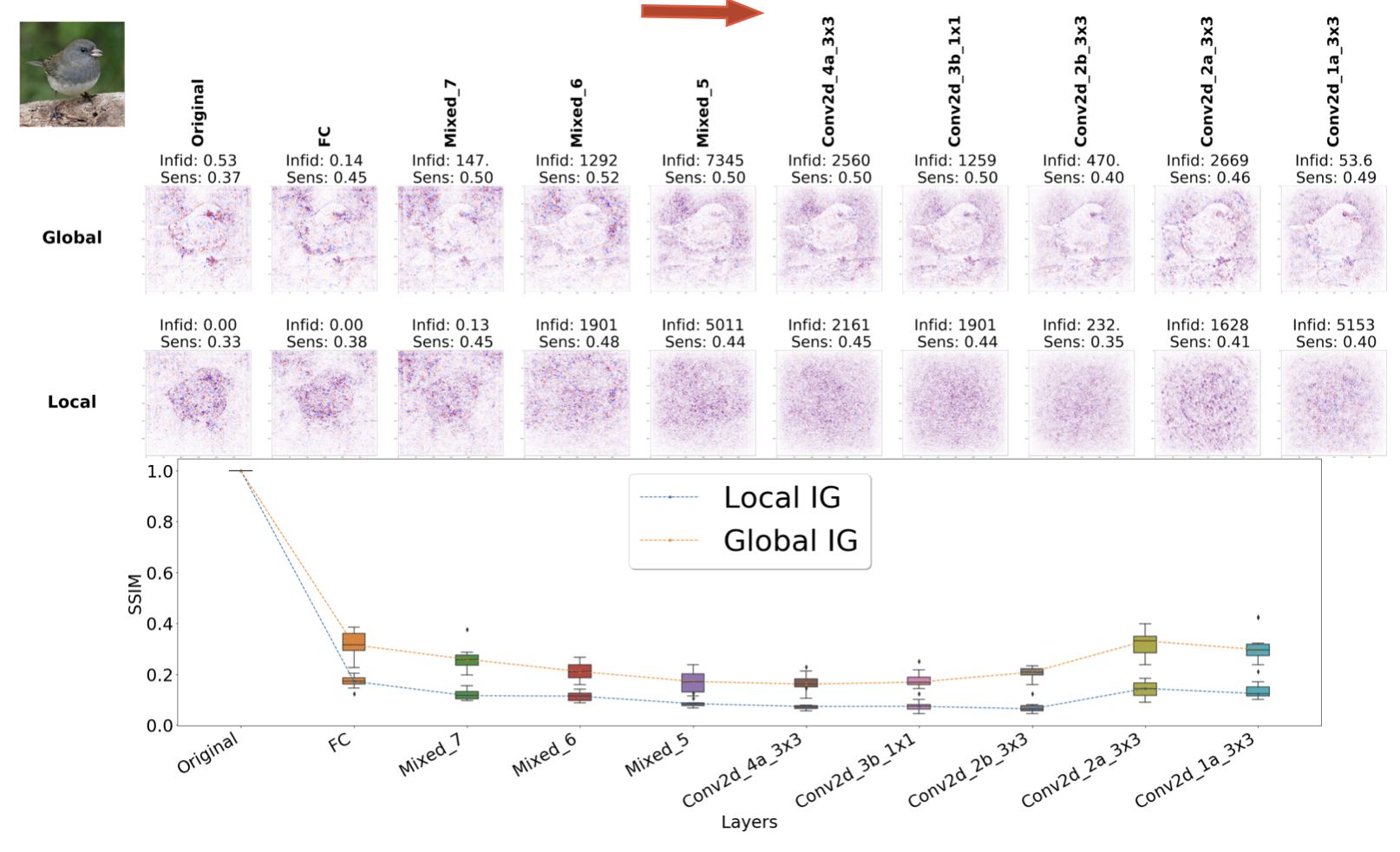}}
\caption{Model explanations (at the top) and SSIM scores (in the bottom) using local and global variants of IG for Inception model when its parameters are randomized in a cascading manner.}
\label{bird_cascading}
\end{center}
\vskip -0.3in
\end{figure}

\begin{figure}[ht]
\begin{center}
\centerline{\includegraphics[width=1.0\columnwidth,scale = 0.01]{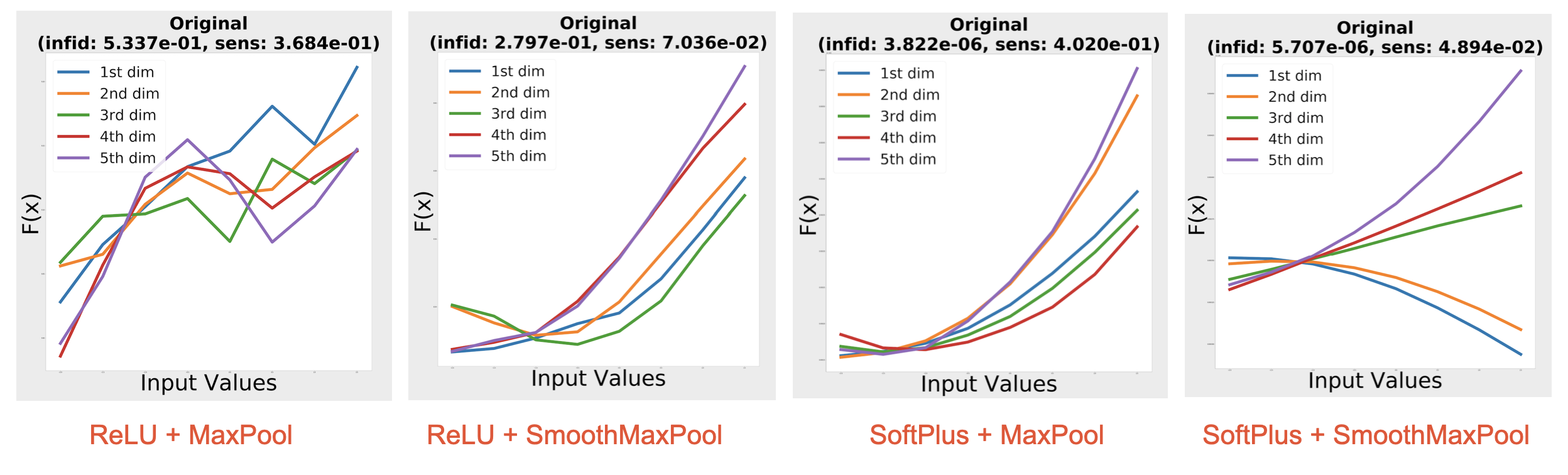}}
\caption{NN function plots for five selected input dimensions with different combinations of smooth and non-smooth activation variants and their relation to infidelity and max-sensitivity metrics.}
\label{smoothing}
\end{center}
\vskip -0.3in
\end{figure}

In our CV experiments we use Inception model and randomize model parameters gradually, layer-wise in a cascading manner, from top to bottom. In Figure~\ref{bird_cascading} we can see that starting from third layer saliency maps look random for the local but not for the global variant. We also observe that the SSIM score for the local variant is almost half of what we observe for the global variant. This indicates how much structural information $x - x_0$ carries. The boxplot distribution of SSIM scores is computed based on 10 random experiments per layer randomization. It is also interesting to observe that the infidelity score increases significantly starting from the second layer, indicating that the explanations are less trustworthy. This is likely because randomizing model parameters makes NN functions less smooth, depending on the chosen randomization technique. For these experiments, we use xavier randomization.
The gradients, in turn, become more noisy and integral approximation becomes less accurate. To alleviate this issue we either increase the number of integral approximation steps or smooth the NN function.
In Figure~\ref{smoothing} we observe that once we replace ReLUs with SoftPlus, the NN function for selected five input dimensions becomes more smooth and infidelity score drops significantly. We observe a similar phenomenon when MaxPool is replaced with a smooth variant such as LSE. Smoothing NN functions also helps to reduce gradient noise and sharp fluctuations, leading to more reliable explanations.

\begin{figure}[ht]
\begin{center}
\centerline{\includegraphics[width=1.0\columnwidth,scale = 0.01]{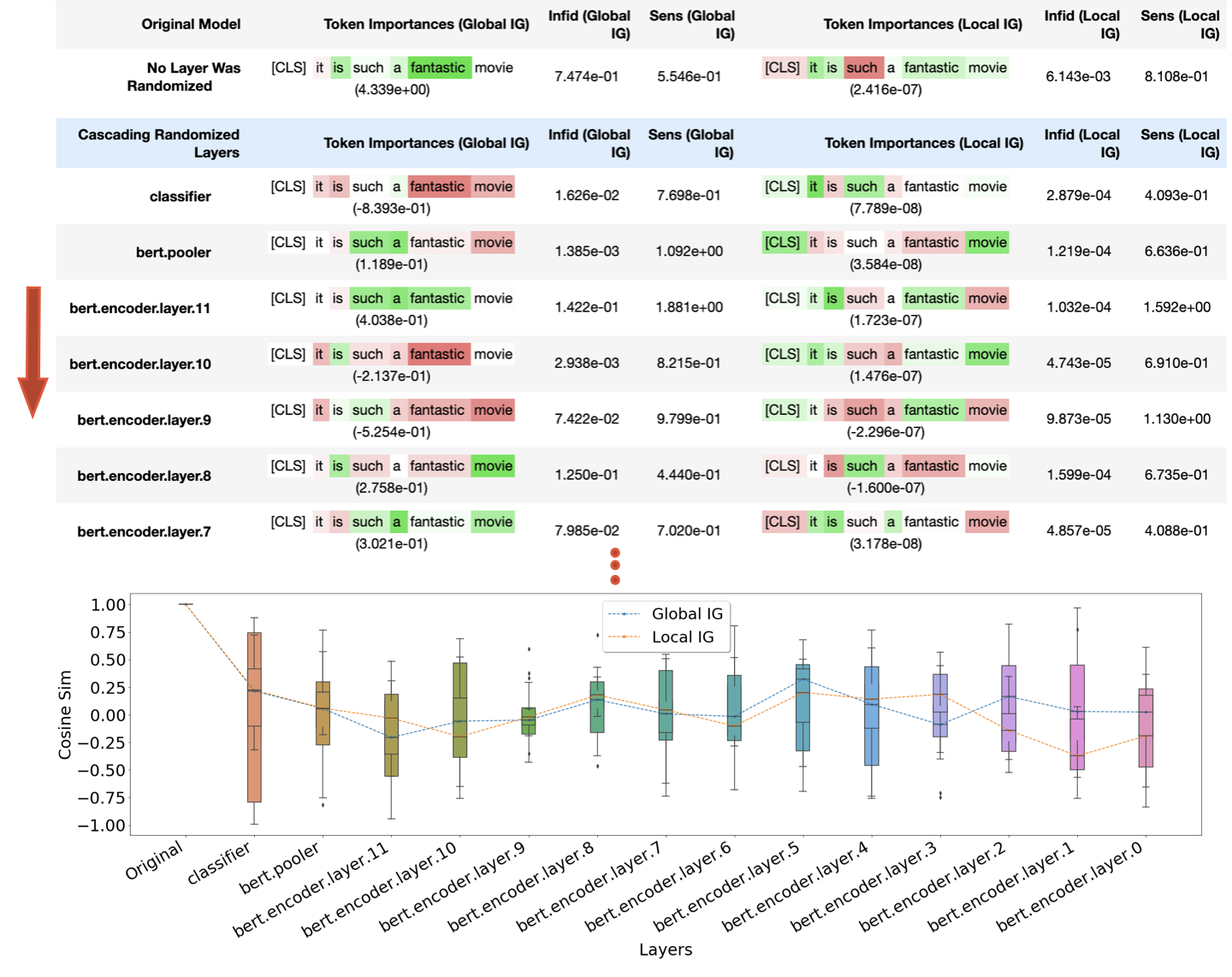}}
\caption{IG explanations and related similarity and evaluation metrics for sentiment analysis model when layer weights are randomized in cascading manner. See Appendix~\ref{sanity_checks_for_text} for more details.}
\label{text_cascading_cosine}
\end{center}
\vskip -0.3in
\end{figure}

For our text experiments we fine-tuned a BERT~\citep{bert} classification model using Hugging Face~\citep{wolf-etal-2020-transformers} with its default parameter setup on SST2 dataset~\citep{sst2} and used $L_2$ norm for normalizing the explanations. Figure~\ref{text_cascading_cosine} visualizes which tokens are correlated with the positive sentiment in green and with the negative in red.

In contrast to the image use case, we note that the global variant of IG does not carry a strong signal of the input structural patterns along the cascading randomization axis. This is related to the fact that a) embedding dimensions do not carry strong structural patterns as images do and b) since token's embedding dimensions are not human-understandable, we summarize importance values along those dimensions. Positive and negative scores cancel each other out, which results in explanations that are more parameter-sensitive. It is also important to observe that both infidelity and max-sensitivity scores remain low or even decrease which indicates that the explanations are more trustworthy and the NN function is more smooth.
In terms of CosSim score, we see that the medians of both variants of IG are close to zero which suggests that the explanations become dissimilar to the original explanations after randomization. In this experiment the distribution of CosSim score was computed for 10 different randomization trials per layer and the perturbations for text are performed in embedding dimension.

In all our experiments the perturbations for infidelity metric and local IG are sampled uniformly from a normal distribution $\displaystyle \sN(0, 0.03)$ and for max-sensitivity from a $L_\infty$ ball with a radius of 0.02. We used five perturbed samples and $L_2$ norm for max-sensitivity. The perturbations for global IG are represented as $x - x_0$. 

\section{Conclusion}
We show that IG returns different qualitative results depending on the model type and application domain. According to our experiments, saliency maps are more sensitive to parameter randomization for text compared with image classification models. We also show that the input multiplier used in many saliency maps carries strong structural patterns especially for images. We recommend that users compare their explanations with and without this multiplier in order to understand the magnitude of these structural effects. We also show that the smoothness of a NN function can affect the infidelity and the quality of explanations and that replacing ReLUs with Softplus and MaxPools with LSE significantly improves the infidelity of explanations. These findings can help model developers to better understand how saliency maps work when implementing Responsible AI frameworks for their models. As a future direction it would be interesting to explore feature pruning tasks and the effects of feature and neuron correlations and interactions on saliency maps.

\bibliography{iclr2021_conference}
\bibliographystyle{iclr2021_conference} 

\appendix
\section{Sanity Checks for Image}
\label{sanity_checks_for_image}
\begin{figure}[ht]
\begin{center}
\centerline{\includegraphics[width=1.0\columnwidth,scale = 0.01]{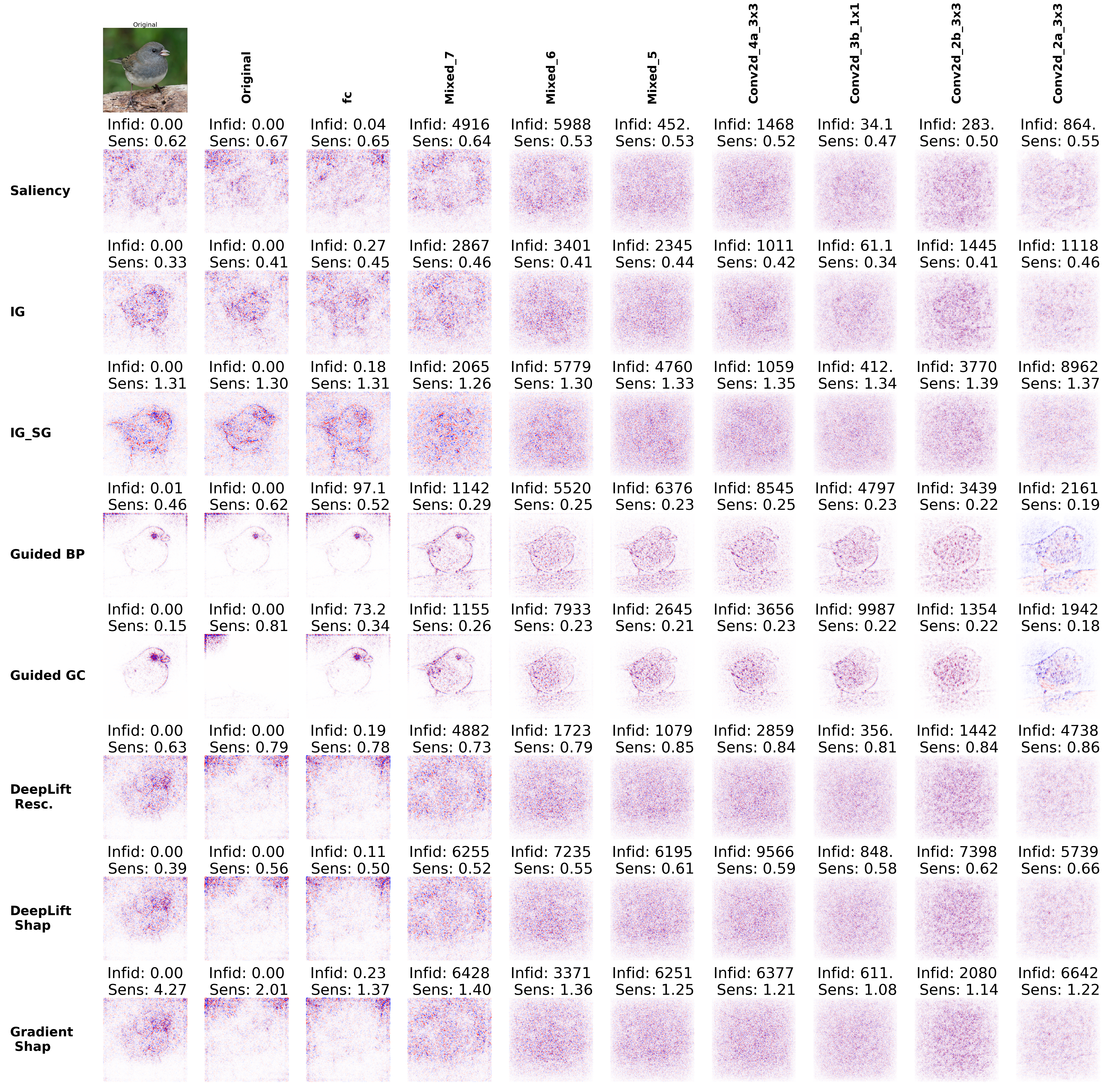}}
\caption{Local explanations, infidelity and max-sensitivity scores for eight different saliency maps when model parameters are randomized in cascading manner.}
\label{bird_explanations_all}
\end{center}
\end{figure}

In this section we describe additional experiments performed for image classification. In Figure~\ref{bird_explanations_all} we visualize saliency maps for eight different local explanation methods. IG, IG w/SmoothGrad (IG\_SG), DeepLift, DeepLift SHAP and Gradient SHAP are computed without $x$ or $x - x_0$ multiplier. As we can see, all methods that do not modify gradients during back-propagation including DeepLift and DeepLift SHAP are sensitive to parameter randomization. We also observe that infidelity score increases significantly starting from the second or third layer of cascading randomization, whereas max-sensitivity remains almost unchanged, for all sensitivity methods. One possible explanation for this could be the fact that parameter randomization makes NN function less smooth which increases the amount of noise in the gradients and infidelity score. Note that this behaviour depends on the randomization technique. In this experiment, we used Xavier random initialization of model weights.

\begin{figure}[ht]
\begin{center}
\centerline{\includegraphics[width=1.0\columnwidth,scale = 0.01]{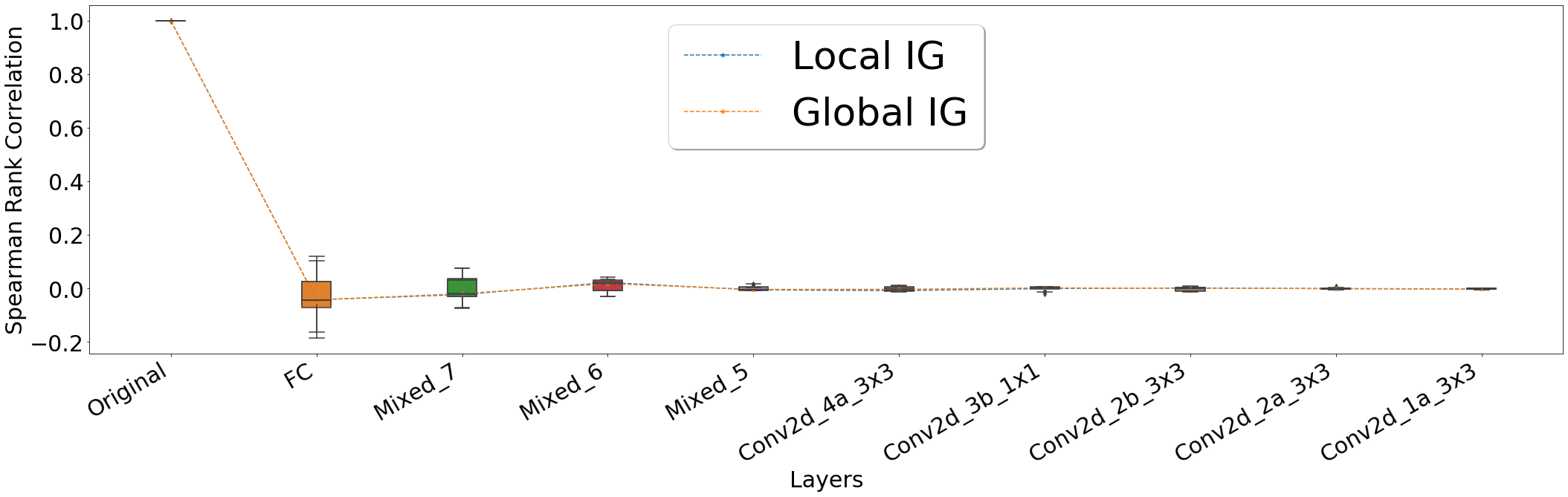}}
\caption{The distribution of Spearman rank correlation scores for local and global explanations of IG with Inception model.}
\label{spearman_image}
\end{center}
\end{figure}

Another way of measuring the quality of our explanations is to look into the correlation of saliency maps before and after randomization. Besides SSIM we also measure Spearman correlation rank both for local and global explanations. Figure~\ref{spearman_image} visualizes Spearman rank-order correlation score when the model is randomized in cascading manner. As also discussed in~\cite{sanity_checks}, Spearman correlation gets close to zero after the first randomization attempt of the last layer and it remains that way until the entire network is fully randomized. We observe that behaviour both for local and global explanations.

\begin{figure}[ht]
\begin{center}
\centerline{\includegraphics[width=1.0\columnwidth,scale = 0.01]{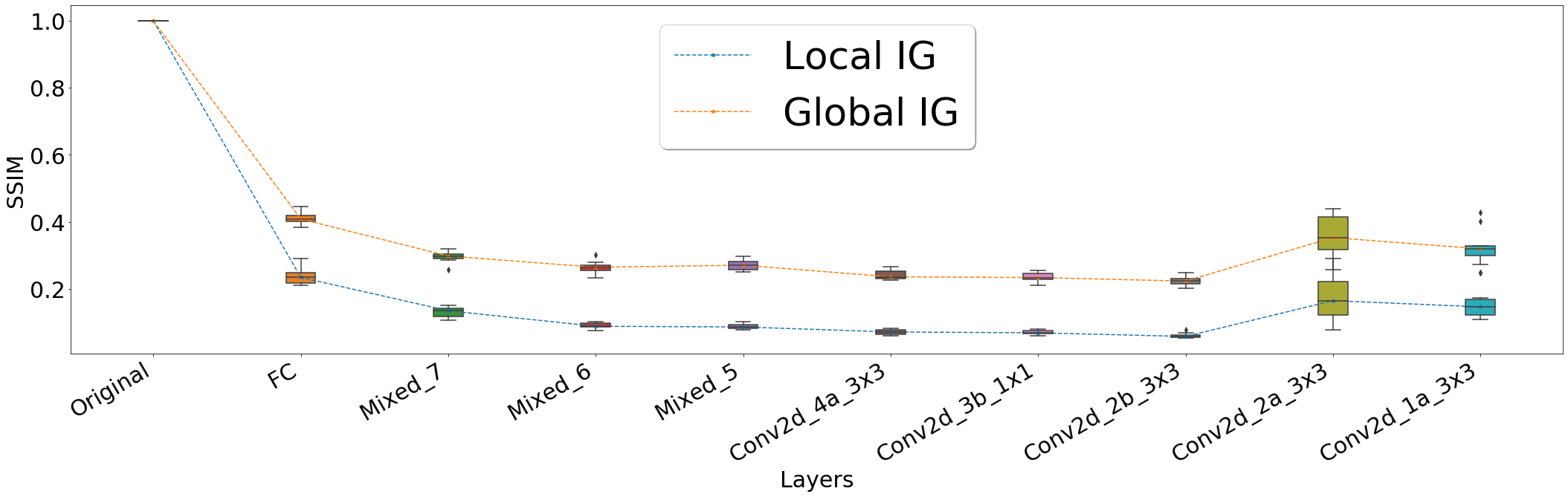}}
\caption{SSIM score averaged across 10 samples for cascading-manner parameter randomization sanity check.}
\label{ssim_10_images}
\end{center}
\end{figure}

In order to convince ourselves that the evaluation and similarity metrics yield similar results for most examples, we run our experiments for a subset of ten examples. We average the infidelity~\ref{infid-sens-10-samples-image}, max-sensitivity~\ref{infid-sens-10-samples-image} and SSIM~\ref{ssim_10_images}
metrics across ten samples and show that their relative distributions are similar to those for one sample.

\begin{table}[t]
\caption{Infidelity and max-sensitivity scores averaged across ten input samples when the layers are randomized in cascading manner for Inception model.}
\label{infid-sens-10-samples-image}
\begin{center}
\begin{tabular}{lllll}
\multicolumn{1}{c}{\bf Layer} & \multicolumn{1}{c}{\bf Infid. Global}  &\multicolumn{1}{c}{\bf Infid. Local} &\multicolumn{1}{c}{\bf Max-Sens. Global} &\multicolumn{1}{c}{\bf Max-Sens. Local}
\\ \hline \\
Original & 2.8433 & 0.0001 & 0.2890 & 0.2621\\
fc & 0.0643 & 1.9918e-06 & 0.2890 & 0.2832\\
Mixed\_7 & 1233.5145 & 0.2014 & 0.3567 & 0.3328\\
Mixed\_6 & 12719875.9301 & 17915.7583 & 0.3835 & 0.3552\\
Mixed\_5 & 826694.3726 & 14482.7274 & 0.3966 & 0.3618\\
Conv2d\_4a\_3x3 & 118765.0183 & 2443.3645 & 0.4428 & 0.3971\\
Conv2d\_3b\_1x1 & 22535.3965 & 1019.3469 & 0.4531 & 0.4084\\
Conv2d\_2b\_3x3 &  6237.2998 & 213.2799 & 0.3865 & 0.3403\\
Conv2d\_2a\_3x3 & 139539.6044 & 3071.9756 & 0.3926 & 0.3580\\
Conv2d\_1a\_3x3 & 1360343.0962 & 3497.0469 & 0.3176 & 0.2951\\
\end{tabular}
\end{center}
\end{table}

\begin{figure}[ht]
\begin{center}
\centerline{\includegraphics[width=1.0\columnwidth,scale = 0.01]{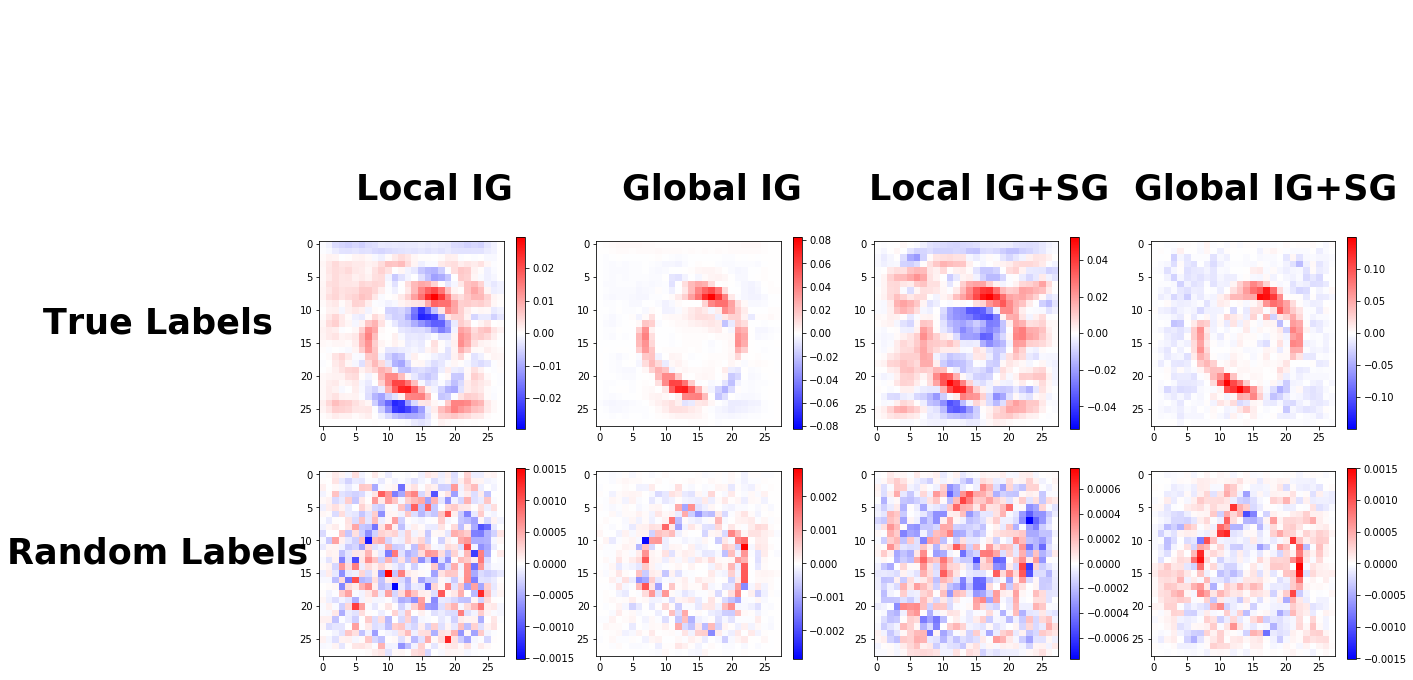}}
\caption{Local and global IG explanations with and without SmoothGrad (SG), before and after label randomization on MNIST dataset.}
\label{label_random_mnist}
\end{center}
\end{figure}

Data randomization is another sanity check that we performed on MNIST dataset. In Figure~\ref{label_random_mnist} we can see how saliency maps change for IG and IG\_SG (with SmoothGrad) for the model that was trained on randomized labels. As we can see, global variants of explanation methods are sensitive to input, whereas local variants carry almost no structural patterns from the input and look random.

\section{Sanity Checks for Text}
\label{sanity_checks_for_text}
\begin{figure}[ht]
\begin{center}
\centerline{\includegraphics[width=1.0\columnwidth,scale = 0.01]{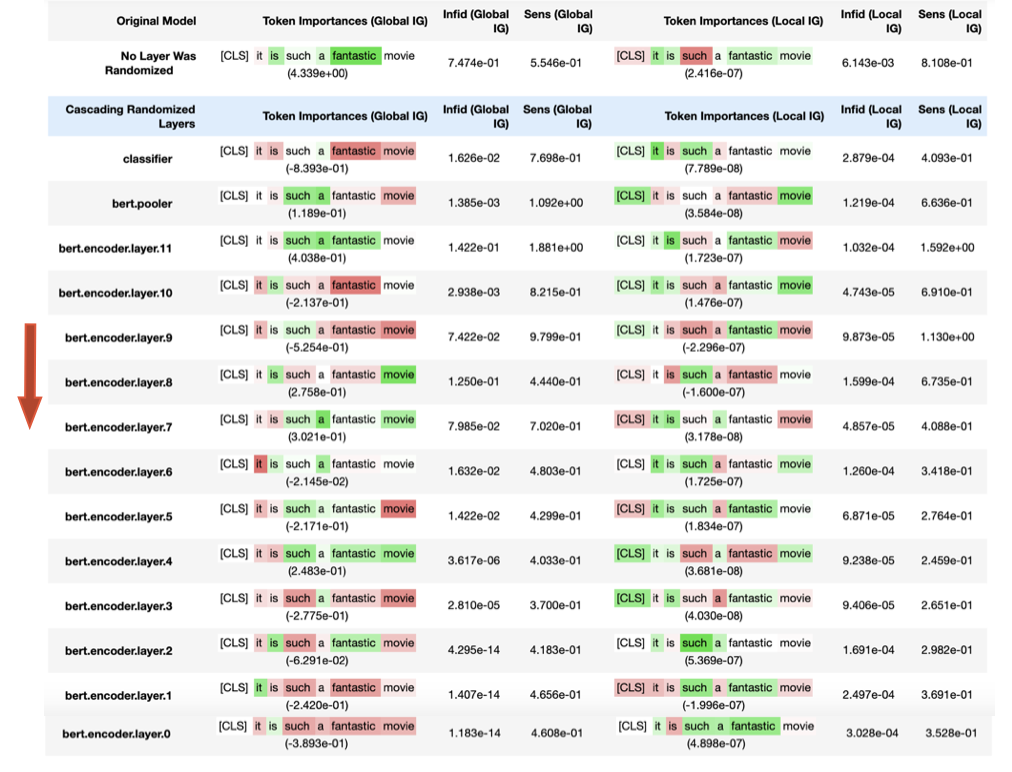}}
\caption{Highlighted local and global IG explanations, corresponding infidelity and max-sensitivity scores and SSIM between the explanations of the original model and the model after randomizing the weights in cascading manner.}
\label{text_cascading_appendix}
\end{center}
\end{figure}

Similar to image modality, we performed data randomization tests for text as well. In Figure~\ref{text_cascading_appendix} we visualize saliency maps for local and global variants of IG for all layers of Bert classification model. We can see that starting from randomization of the first layer the saliency maps look arbitrary and do not carry structural patters from input both for global and local variants of IG.

\begin{figure}[ht]
\begin{center}
\centerline{\includegraphics[width=1.0\columnwidth,scale = 0.01]{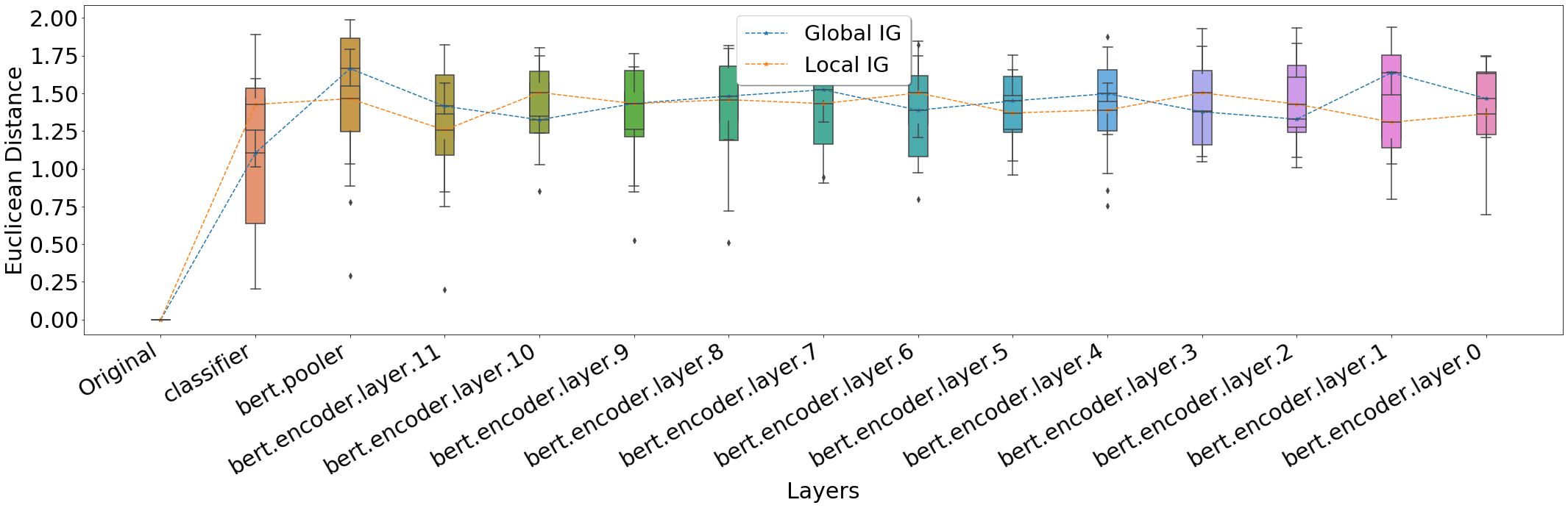}}
\caption{Euclidean distance between explanations for the original model and after randomizing the weights in cascading manner.}
\label{text_cascading_euclidean}
\end{center}
\end{figure}

In addition to cosine similarities, we also compute euclidean distances between the explanation maps for the original model and after parameter randomization. In Figure~\ref{text_cascading_euclidean} we can see that explanations become dissimilar right after randomizing the first layer and that both global and local variants of IG have dissimilarity patterns.

\begin{figure}[ht]
\begin{center}
\centerline{\includegraphics[width=1.0\columnwidth,scale = 0.01]{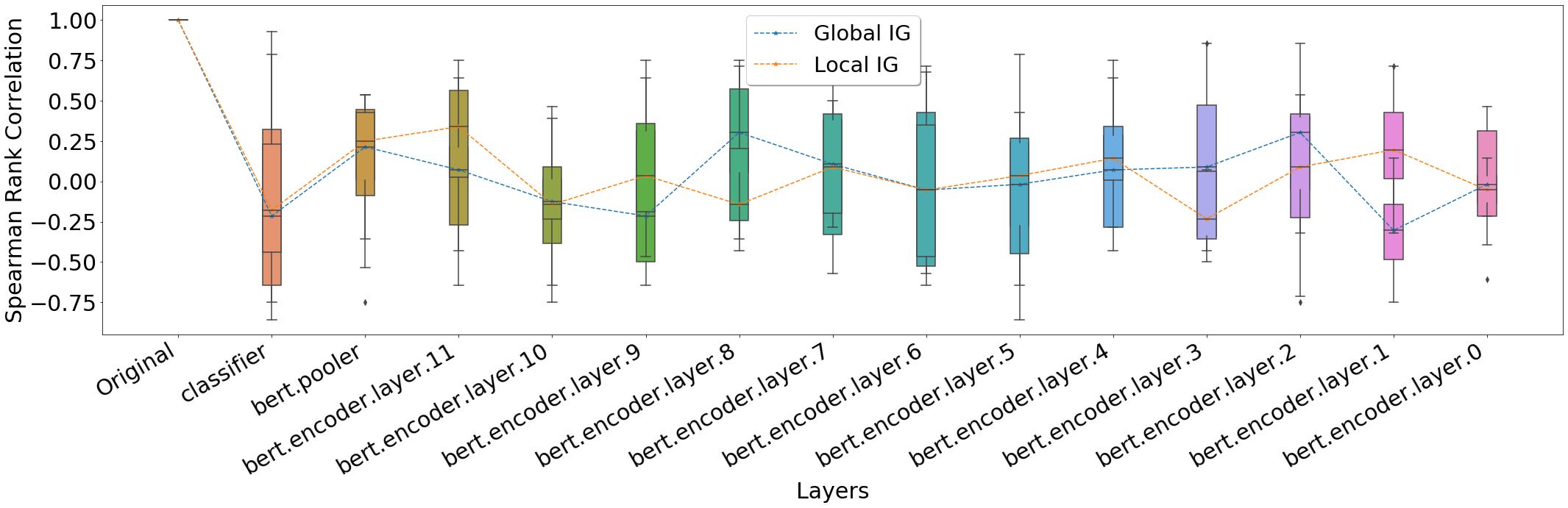}}
\caption{The distribution of Spearman rank correlation scores for local and global explanations of IG with Sentiment Analysis Bert model.}
\label{spearman_text}
\end{center}
\end{figure}

Furthermore, we computed Spearman rank correlation between saliency maps of local and global explanations~\ref{spearman_text}. In contrary to image, the variance of correlation scores is larger here, however the median is still close to zero. This means that after randomization saliency maps become fairly dissimilar.

\begin{figure}[ht]
\begin{center}
\centerline{\includegraphics[width=1.0\columnwidth,scale = 0.01]{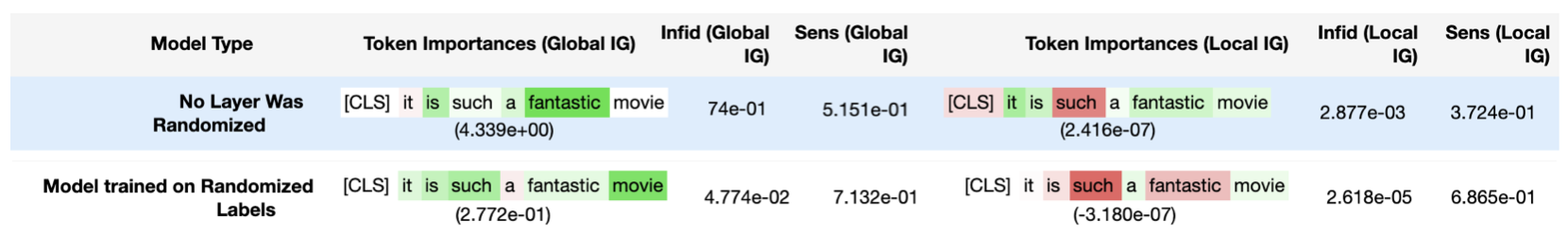}}
\caption{Model explanations for the original model and the model that was trained on randomly shuffled data labels.}
\label{text_data_random}
\end{center}
\end{figure}

In addition to that, we performed data randomization sanity check for text as well. In Figure~\ref{text_data_random} we can see highlighted explanations using IG for the original model and the model that was trained using randomized data labels. As we can see the explanations for those too models look pretty different and the quality of explanations doesn't seem to be changed much based on infidelity and max-sensitivity metrics.

\begin{figure}[ht]
\begin{center}
\centerline{\includegraphics[width=1.0\columnwidth,scale = 0.01]{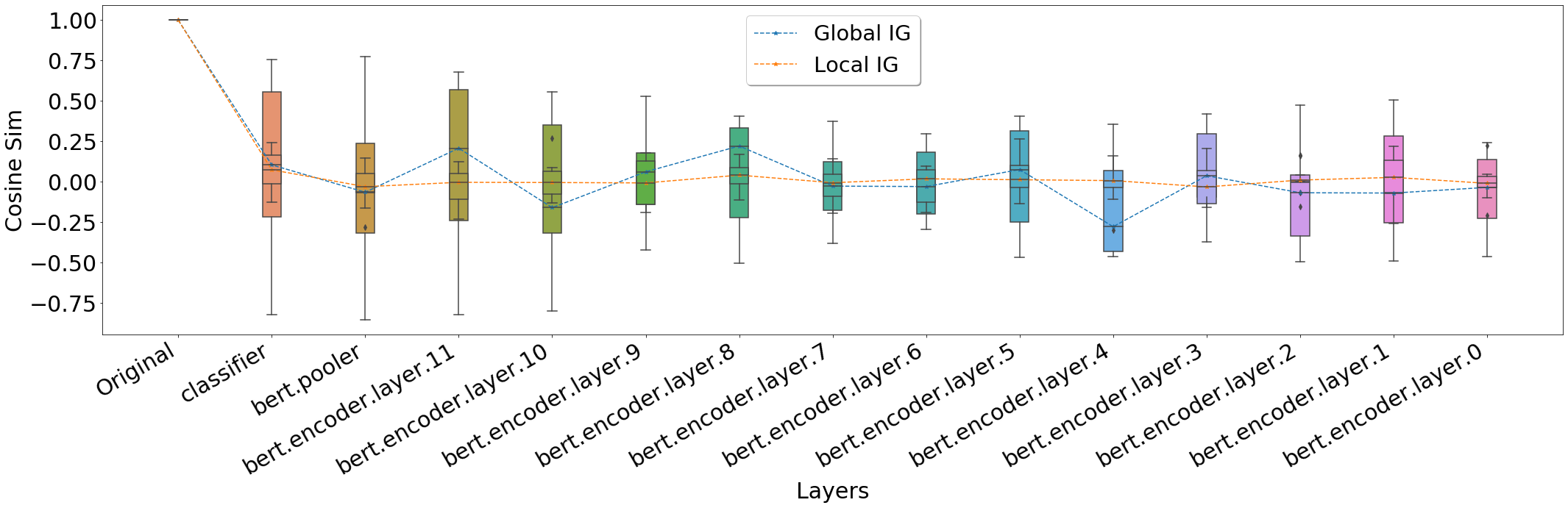}}
\caption{Cosine similarity score averaged across 10 samples for cascading-manner parameter randomization sanity check.}
\label{cossim_10_sentences}
\end{center}
\end{figure}

Similar to image, we would like to examine how the evaluation and similarity metrics change for a subset of ten samples. Figures~\ref{infid-sens-10-samples-text} and ~\ref{cossim_10_sentences} visualize the distribution of infidelity, max-sensitivity and cosine similarity metrics. As we can see, all those metrics change with the patterns similar to those that we obtained for one sample.

\begin{table}[t]
\caption{Infidelity and max-sensitivity scores averaged across ten input samples when the layers are randomized in cascading manner for Bert Classification model.}
\label{infid-sens-10-samples-text}
\begin{center}
\begin{tabular}{lllll}
\multicolumn{1}{c}{\bf Layer} & \multicolumn{1}{c}{\bf Infid. Global}  &\multicolumn{1}{c}{\bf Infid. Local} &\multicolumn{1}{c}{\bf Max-Sens. Global} &\multicolumn{1}{c}{\bf Max-Sens. Local}
\\ \hline \\
Original & 8.63892e-17 & 4.70457e-17 & 0.33020 & 0.03452\\
classifier & 6.49653e-17 & 7.19617e-17 & 0.31056 & 0.03374\\
bert.pooler & 1.56992e-17 & 1.94419e-17 & 0.31914 & 0.03371\\
bert.encoder.layer.11 & 3.76434e-17 & 4.45455e-17 & 0.33087 & 0.03442\\
bert.encoder.layer.10 & 9.04224e-18 & 1.719544e-17 & 0.31933 & 0.03463\\
bert.encoder.layer.9 & 4.40619e-17 & 2.42904e-17 & 0.31151 & 0.03369\\
bert.encoder.layer.8 & 3.63641e-17 & 3.348883e-17 & 0.33557 & 0.03355\\
bert.encoder.layer.7 & 1.11239e-17 & 4.37865e-17 & 0.30347 & 0.03474\\
bert.encoder.layer.6 & 3.40005e-17 & 4.78741e-17 & 0.30498 & 0.03478\\
bert.encoder.layer.5 & 2.25947e-17 & 3.35171e-17 & 0.29744 & 0.03399\\
bert.encoder.layer.4 & 1.98625e-17 & 3.09084e-17 & 0.29910 & 0.03434\\
bert.encoder.layer.3 & 3.530162e-17 & 4.18133e-17 & 0.2714 & 0.03469\\
bert.encoder.layer.2 & 1.74339e-17 & 3.06048e-17 &  0.31196 & 0.03423\\
bert.encoder.layer.1 & 7.82793e-18 & 2.21003e-17 & 0.28099 & 0.03310\\
\end{tabular}
\end{center}
\end{table}

\end{document}